\documentclass[letterpaper]{article}
\usepackage{aaai21}
\usepackage{flushend}
\usepackage{times}
\usepackage{helvet}
\usepackage{courier}
\usepackage{graphicx}
\usepackage{subfigure}
\usepackage{multirow}
\usepackage{xcolor}
\usepackage{natbib}
\usepackage{booktabs} 
\usepackage{amsmath,amsfonts,bm}
\usepackage{url}
\usepackage{comment}
\usepackage{xspace}
\usepackage[switch]{lineno}

\begin{document}

%
\title{Learning from History: Modeling Temporal Knowledge Graphs with Sequential Copy-Generation Networks}

\author{Cunchao Zhu\textsuperscript{\rm 1}, Muhao Chen\textsuperscript{\rm 2}, Changjun Fan\textsuperscript{\rm 1}, Guangquan Cheng\textsuperscript{\rm 1}, Yan Zhang\textsuperscript{\rm 3}\\
}

\affiliations {
    \textsuperscript{\rm 1} College of Systems Engineering, National University of Defense Technology, China \\
    \textsuperscript{\rm 2} Viterbi School of Engineering, University of Southern California, USA \\
    \textsuperscript{\rm 3} École Pour l'Informatique et les Techniques Avancées, France \\    
    \texttt{\{zhucunchao19,fanchangjun,cgq299\}@nudt.edu.cn, muhaochen@usc.edu, yan.zhang@epita.fr}
}


\newcommand{\model}[0]{\texttt{CyGNet}\xspace}
\newcommand{\FC}[1]{{\color{red}[{\sc FC:} #1]}}
\newcommand{\muhao}[1]{{\color{blue}[{\sc MC:} #1]}}
\newcommand{\ZC}[1]{{\color[RGB]{148,0,211}[{\sc ZC:} #1]}}
\newcommand{\todo}[1]{{\bf\color{red}[{TODO:} #1]}}
\newcommand{\nop}[1]{}
\newcommand{\stitle}[1]{\vspace{1ex} \noindent{\bf #1}}

\maketitle


\begin{abstract}
\begin{quote}

\nop{
\FC{please keep the following schema in mind to organize your abstract}

\FC{background}

\FC{limitation and challenge}

\FC{solution}

\FC{results}
}

Large knowledge graphs often grow to store temporal facts that model the dynamic relations or interactions of entities along the timeline. Since such temporal knowledge graphs often suffer from incompleteness, it is important to develop time-aware representation learning models that help to infer the missing temporal facts.
While the temporal facts are typically evolving,
it is observed that many facts often show a repeated pattern along the timeline, such as economic crises and diplomatic activities.
This observation indicates that a model could potentially learn much from the known facts appeared in history.
To this end, we propose a new 
representation learning model for temporal knowledge graphs, namely \model \includegraphics[height=1em]{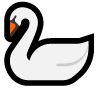},
based on a novel time-aware copy-generation mechanism.
\model is not only able to predict future facts from the whole entity vocabulary, but also capable of identifying facts with repetition and accordingly predicting such future facts with reference to the known facts in the past.
We evaluate the proposed method on the knowledge graph completion task using five benchmark datasets. Extensive experiments demonstrate the effectiveness of \model for predicting future facts with repetition as well as 
\emph{de novo} fact prediction.
\end{quote}
\end{abstract}

\section{Introduction}

\begin{figure}[!t]
\centering
\includegraphics[width=0.48\textwidth]{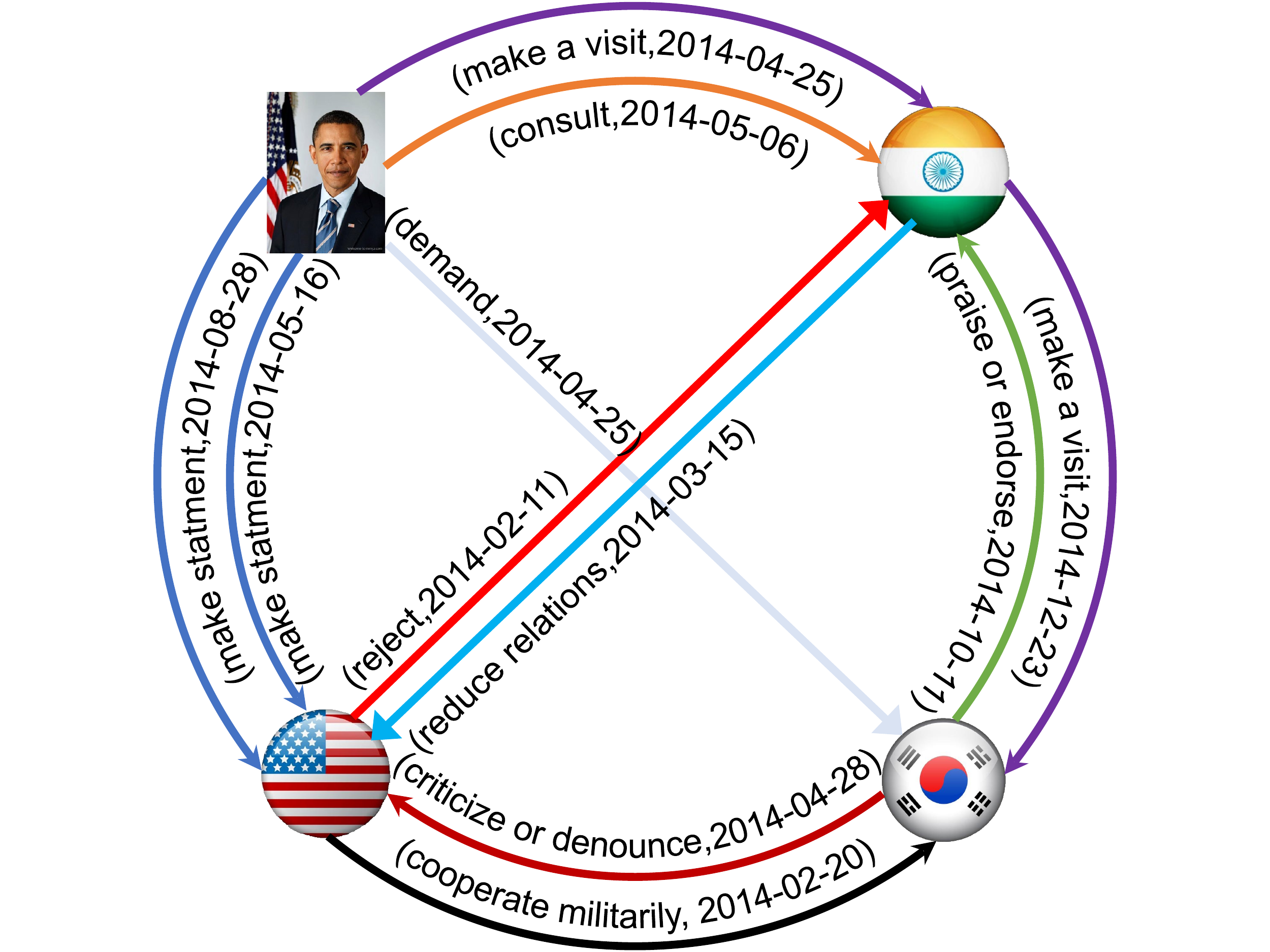}
\vspace{-1em}
\caption{A snippet of ICEWS showing several records of diplomatic activities.
}\label{snap-shot example of TKG}
\end{figure}

Knowledge Graphs (KGs) are widely used resources for knowledge representations of real-world facts (or events),
inasmuch as it supports countless knowledge-driven tasks in areas such as information retrieval \citep{liu2018entity}, natural language understanding \citep{chen2019incorporating,he2017learning},
recommender systems \citep{koren2009matrix} and healthcare \cite{hao2020bio}.
Traditionally, a KG only possesses facts in a static snapshot, while currently the rapid growing data often exhibit complex temporal dynamics.
This calls for new approaches to model such dynamics of facts by assigning the interactions of entities with temporal properties (i.e., known as \emph{temporal knowledge graphs}, or TKGs.). Representative TKGs include Global Database of Events, Language, and Tone (GDELT) \citep{leetaru2013gdelt} and Integrated Crisis Early Warning System (ICEWS) \citep{boschee2015icews}, which are two well-known event-based data repository that store evolving knowledge about entity interactions across the globe. Figure~\ref{snap-shot example of TKG} shows a snippet of ICEWS, 
which shows several records of diplomatic activities for different timestamps.

\nop{
As the structured knowledge only holds within a specific period and the evolution of facts follows a time sequence, it is necessary to model such temporal dynamics of facts by assigning the interactions of entities with temporal properties. Accordingly, KGs are extended to TKGs.
}


\nop{
For instance, Global Database of Events, Language, and Tone (GDELT) \citep{leetaru2013gdelt} and Integrated Crisis Early Warning System (ICEWS) \citep{boschee2015icews} are two popular event based data repository that can store such temporal information, e.g. $(Barack\;Obama,Make\;a\;visit,Ukraine,2014-07-08)$.}


Recently, many research efforts have put into representation learning for TKGs. Relevant methods typically encode the temporally evolving facts of entity relations with time-specific embeddings. This provides a versatile and efficient tool to complete future facts of TKGs based on the embedding representations of the past facts. Moreover, it benefits a wide range of downstream applications, e.g. transaction recommendation \citep{lortscher2010system}, event process induction \citep{bouchard2012temporal,du2016recurrent} and social relation prediction \citep{zhou2018dynamic}. Several temporal knowledge graph embedding (TKGE) methods only focus on calculating latent representations for each snapshot separately, and thus are not capable of capturing the long-term dependency of facts in consecutive time snapshots \citep{jiang2016towards,dasgupta2018hyte,wang2019hybrid,ma2019embedding,lacroix2019tensor}.
Some other recent attempts encode the temporally evolving facts of entity relations by incorporating past information from previous snapshots \citep{goel2019diachronic,trivedi2017know,jin2020Renet}.
However, the complex evolution of temporal facts, including 
recurrence and trending patterns, can cause the aforementioned methods to fall short.

\nop{
Extensive experiments have shown that TKGE models 
outperform traditional KGE models \citep{bordes2013translating,yang2014embedding,trouillon2016complex,kazemi2018simple} on link prediction over temporal KGs. 
\muhao{$<-$ This claim is trivial.}
Some TKGE models \citep{jiang2016towards,dasgupta2018hyte,wang2019hybrid,garcia2018learning,goel2019diachronic,lacroix2019tensor} focus on extending the KG representation method of static graphs to utilize time, such as calculating a hidden representation for each timestamp, and extending the score function to use timestamp representations and entity and relationship representations. However, these models cannot predict future events \muhao{you haven't ever defined events. You are dealing with entities. You need to be careful, NLP people have a strict definition of events. You don't want to raise confusion.}, as representations of unseen time stamps are unavailable. \citeauthor{jin2020Renet} uses a recurrent event encoder to predict future event. \muhao{In this paragraph you need to have a representative enough summarization of the previous methods. You need to let people believe you have covered basically all kinds of methods (instead of just some methods) in a concise paragraph, and those methods commonly fall short due to one or two reasons.}
}

In fact, many facts occur repeatedly along the history. For instance, global economic crises happen periodically in about every seven to ten years \citep{korotayev2010spectral}; 
the diplomatic activities take place regularly between two countries with established relationships \citep{feltham2004diplomatic};
East African animals undergo annual vast migration in every June \citep{musiega2004simulating}. More specifically, 
we found that over 80\% of all the events throughout the 24 years of ICEWS data (i.e. 1995 to 2019) have already appeared during the previous time period. This phenomenon highlights the importance of leveraging the known facts in predicting the future ones. 
However, most existing methods do not incorporate the awareness of such evolution patterns in modeling a TKG.
On the contrary, to conform with the nature of evolving facts in TKG events, we believe more precise temporal fact inference should make full use of known facts in history. Accordingly, a TKGE model would benefit from learning the temporal recurrence patterns while characterizing the TKGs.

\begin{figure*}[t!]
\centering
\includegraphics[width=0.9\textwidth]{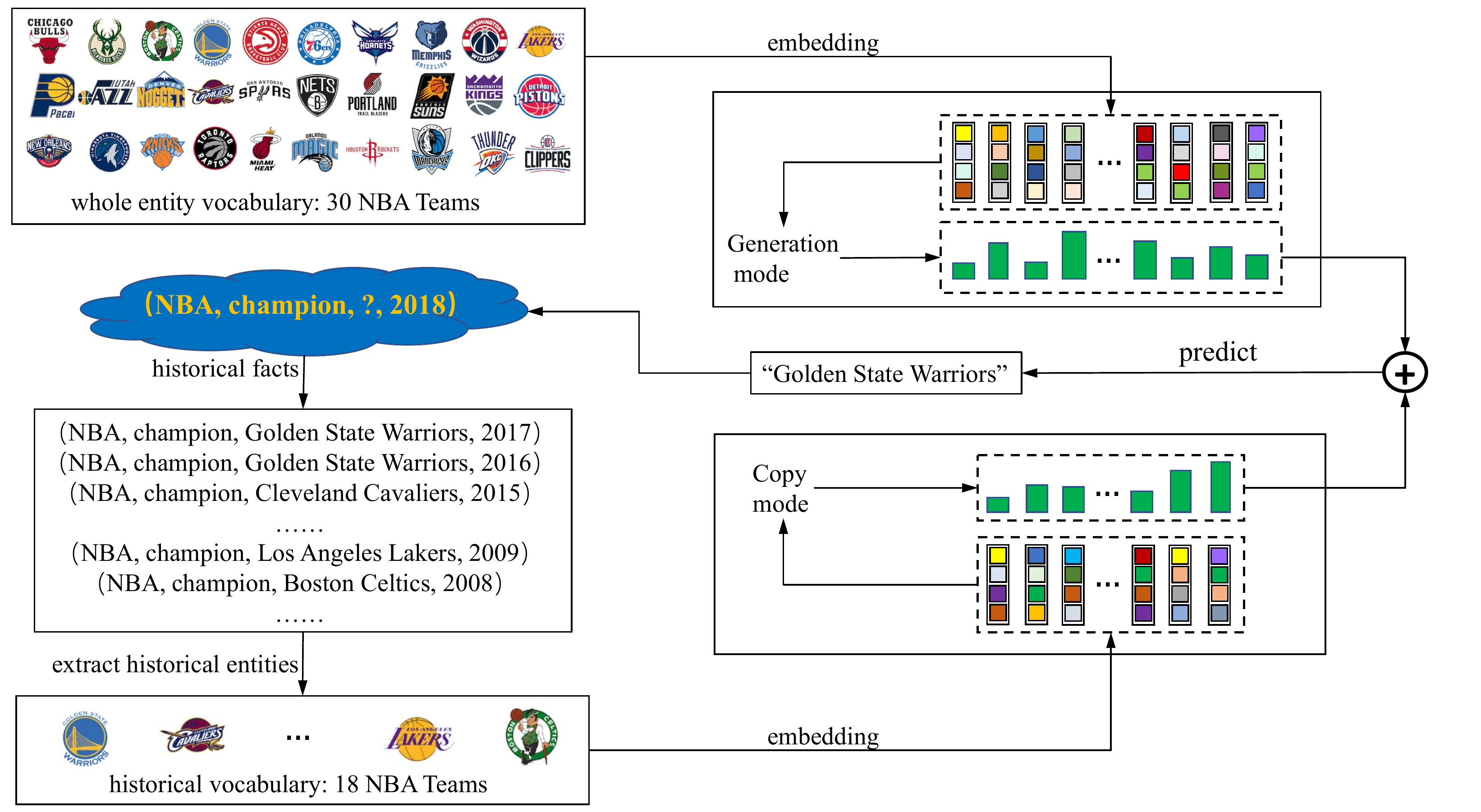} 
\caption{Illustration of \model's copy-generation mechanism. In this figure, when predicting which team was the champion in NBA in 2018. 
\model first obtains each entity's embedding vector (shown as a color bar). It then utilizes both inference modes to infer entity probabilities (shown as green bars, with heights proportional to the probability values) from the historical or the whole entity vocabulary.
''\emph{Golden State Warriors}'' is predicted as the 2018 NBA champion by combining probabilities from both Copy and Generation modes.
}\label{fig:intro_sample}
\end{figure*}

To this end, we propose a new representation learning method for TKGs based on a novel time-aware copy mechanism. 
The proposed \model\includegraphics[height=1em]{figs/1f9a2.png} 
(Temporal \textbf{\underline{C}}op\textbf{\underline{y}}-\textbf{\underline{G}}eneration \textbf{\underline{Net}}work)
is not only able to predict future facts from the whole entity vocabulary, but also capable of identifying facts with recurrence, and accordingly selecting such facts based on the historical vocabulary of entities that form facts only appeared in the past.
This behaves similarly to the copy mechanism in the abstractive summarization task \citep{gu2016incorporating} in natural language generation (NLG), which allows a language generator to choose to copy subsequences from the source text, so as to help generate summaries that preserve salient information in the source text.
Inspired by this mechanism, when predicting a future fact of quadruple $(s_{i},p_{j},?,t_{T})$, we can treat the known facts $\{(s_{i},p_{j},o_{a},t_{0}),(s_{i},p_{j},o_{b},t_{0}),...,(s_{i},p_{j},o_{k},t_{T-1})\}$ appeared in the previous snapshots as the source text in abstractive summarization, and predict the future facts based solely on the known facts from the historical vocabulary $\{o_a,o_b,...,o_k\}$.
As shown in Figure~\ref{fig:intro_sample}, \model consists of two modes of inference, namely Copy mode and Generation mode. 
We illustrate \model's two inference modes with an example prediction of the 2018 NBA championship. 
Accordingly, all 18 NBA champion teams before 2018 are collected as the historical entity vocabulary, and the total 30 NBA teams are considered as the whole entity vocabulary. To complete the query $($\emph{NBA},\emph{champion},\emph{?},2018$)$, \model utilizes the Copy mode to predict the entity probabilities from the known entity vocabulary, and adopts the Generation mode to infer the entity probabilities from the whole entity vocabulary. Then, both probability distributions are combined as the final prediction.

We conduct extensive experiments on five benchmark TKG datasets, which demonstrate that \model can effectively model TKG data via the combination of the Copy mode and the Generation mode in both learning and inference.
Accordingly, \model achieves more precise future fact predictions in most datasets than state-of-the-art (SOTA) methods that do not take special consideration of temporal facts with recurrence patterns. 

This work presents the following contributions:




\begin{enumerate}
    \setlength\itemsep{1pt}
    \item We investigate the underlying phenomena of temporal facts with recurrence, and propose to make reference to known facts in history when learning to infer future facts in TKGs;
    \item We propose a novel TKGE model \model via the time-aware copy-generation mechanism, which combines two modes of inference to make predictions based on either the historical vocabulary or the whole entity vocabulary, hence being more consistent to the aforementioned evolution pattern of TKG facts; 
    \item We conduct extensive experiments on five public TKG benchmarks datasets, and demonstrate its effectiveness in future fact (link) prediction. 
\end{enumerate}



\section{Related Work}
\label{sec:related_work}
We discuss three lines of relevant research. Each has a large body of work, of which we can only 
provide a highly selected summary.

\stitle{Static KG Embeddings} A large number of methods are developped to model static KGs without temporally dynamic facts, which have been summarize in recent surveys \citep{wang2017knowledge,ji2020survey,dai2020survey}. A class of these methods are the translational models \citep{bordes2013translating,wang2014knowledge,ji2015knowledge}, which models a relation between two entity vectors as a translation vector. Another class of models are the semantic matching models that measures plausibility of facts using a triangular norm \citep{yang2014embedding,trouillon2016complex,sun2019rotate}.
Some other models are based on deep neural network approaches using feed-forward or convolutional layers on top of the embeddings \citep{schlichtkrull2018modeling,dettmers2018convolutional,schlichtkrull2018modeling}.
However, these methods do not capture temporally dynamic facts.

\stitle{Temporal KG Embeddings} More recent attempts have been made to model the evolving facts in TKGs. TTransE \citep{jiang2016towards} is an extension of TransE \citep{bordes2013translating} by embedding temporal information into the score function. HyTE \citep{dasgupta2018hyte} replaces the unit normal vector of the hyperplane projection in TransH \citep{wang2014knowledge} with a time-specific normal vector. Know-Evolve \citep{trivedi2017know} learns non-linearly evolving entity representations over time which models the occurrence of a fact as a temporal point process. DE-SimplE \citep{goel2019diachronic} leverages diachronic embeddings to represent entities at different timestamps and employs the same score function as SimplE \citep{kazemi2018simple} to score the plausibility of a quadruple. Based on the Tucker decomposition \citep{balazevic2019tucker}, ConT \citep{ma2019embedding} learns a new core tensor for each timestamp. 
However, these models do not provide a mechanism to capture the long-term dependency of facts in consecutive time snapshots. 

Some other methods are designed to model graph sequences, which can be applied to capture long-term dependency of TKG facts. 
TA-DistMult \citep{garcia2018learning} utilizes a recurrent neural network to learn time-aware representations of relations and uses standard scoring functions from DistMult \cite{yang2014embedding}. 
GCRN \citep{seo2018structured} combines GCN \citep{kipf2017semi} for graph-structured data and RNN to identify simultaneously meaningful spatial structures and dynamic patterns.
DyREP \citep{trivedi2018dyrep} divides the dynamic graph network into two separate processes of global and local topological evolution, and proposes a two-time scale deep temporal point process to jointly model the two processes. 
Know-Evolve, DyREP and GCRN have also been combined with MLP decoders to predict future facts, as presented by \citet{jin2020Renet}. The previous SOTA method in this line of research, i.e. RE-NET \citep{jin2020Renet} models event (fact) sequences jointly with an RNN-based event encoder and an RGCN-based \cite{schlichtkrull2018modeling} snapshot graph encoder.

\stitle{Copy Mechanism} The copy mechanism has been previously applied to NLG tasks, particularly for abstractive summarization. \citet{vinyals2015pointer} proposed the Pointer Networks in which a pointer mechanism 
is used to select the output sequence directly from the input sequence. However, the Pointer Network cannot make prediction using lexemes that are external to the input sequence. COPYNET \citep{gu2016incorporating} solves this problem in a hybrid end-to-end model by combining the pointer network (or termed as copy mechanism in its context) with a generation mechanism that yields lexemes that do not appear in the input. SeqCopyNet \citep{zhou2018sequential} improves the copy mechanism to not only copy single lexemes, but also copies subsequences from the input text. Our work is inspired to use the copy mechanism for the characteristics of TKGs, which to the best of our knowledge, is the first work that incorporates the copy mechanism into modeling TKGs. This is consistent to the nature that temporal knowledge may contain recurrence patterns along the trending timeline.
\section{Method}
\label{sec:model}

In this section, we introduce the proposed model, named \model, for fact/link prediction in TKGs. We start with the notations, and then introduce the model architecture as well as its training and inference procedures in detail.







\begin{figure}[!t]
\centering
\includegraphics[width=0.47\textwidth]{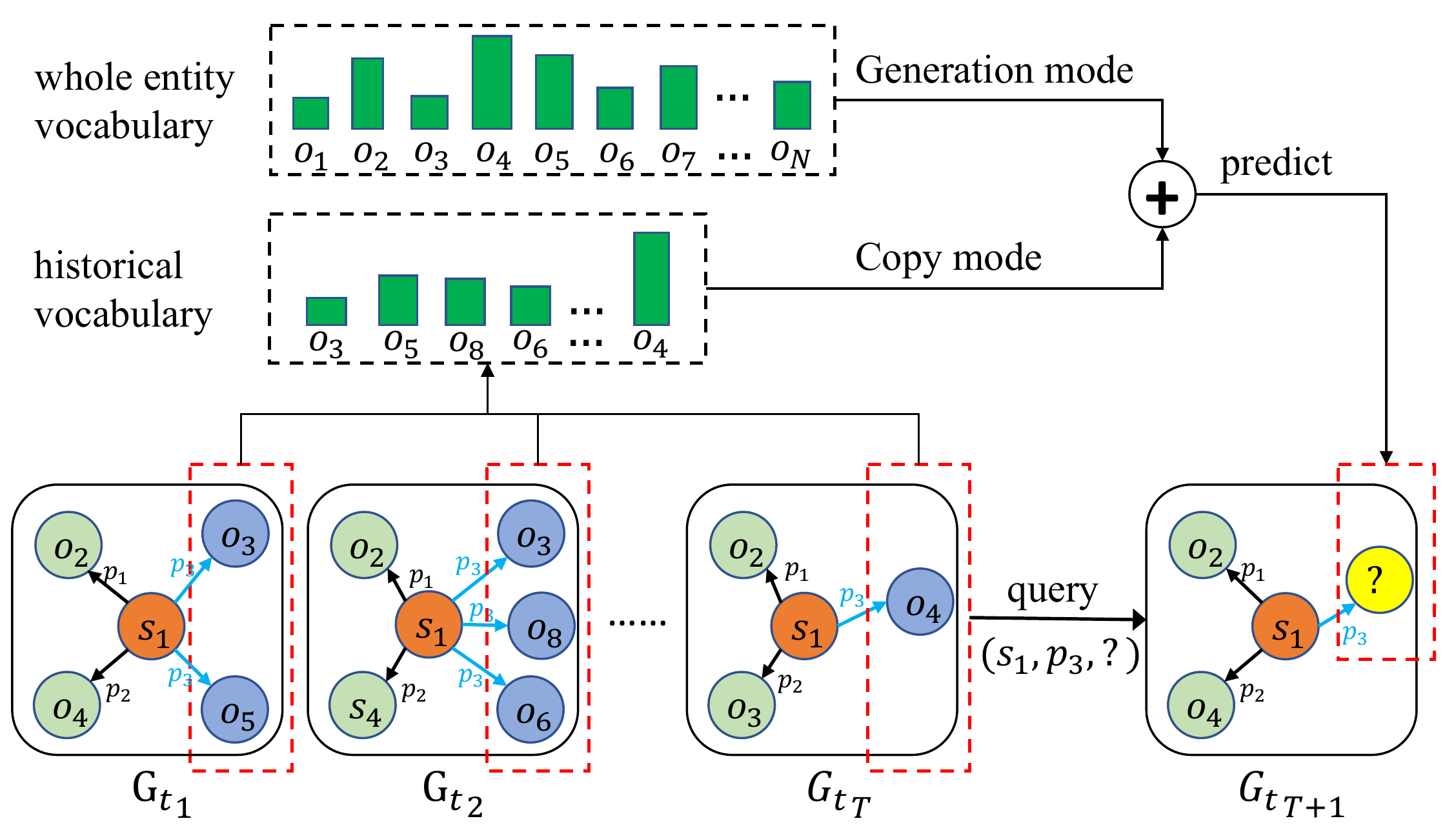} 
\caption{The overall architecture of \model. 
Dark blue nodes are the candidate object entities in the historical vocabulary for query $(s_1,p_3,?,t_{T+1})$. Green bars (with heights proportional) indicate the probability values predicted by the Copy mode and the Generation mode.}\label{fig:model}
\end{figure}

\subsection{Notations}

TKGs incorporate temporal information in traditional KGs. In a TKG, each of the fact captures the relation (or predicate) $p\in \mathcal{R}$ for subject and object entity $s\in \mathcal{E}$ and $o\in \mathcal{E}$ at time step $t\in \mathcal{T}$, where $\mathcal{E}$, $\mathcal{R}$ denote the corresponding vocabularies of entities and relations respectively, and $\mathcal{T}$ is the set of timestamps.
Boldfaced $\mathbf{s}$, $\mathbf{p}$, $\mathbf{o}$, $\mathbf{t}$ represent the embedding vectors of subject entity $h$, predicate $p$, object entity $o$ and time step $t$ in a temporal fact.
$\mathcal{G}_t$ is a snapshot of the TKG at a time $t$, and $g=(s,p,o,t)$ denotes a quadruple (fact) in $\mathcal{G}_t$. A TKG is built upon a set of fact quadruples ordered ascendingly based on their timestamps, i.e., $\mathcal{G}=\{\mathcal{G}_1,\mathcal{G}_2,…,\mathcal{G}_T\}$, where the same quadruple is removed for redundancy. 
For each subject entity and predicate pair at time step $t_k$, we define  a delimited subset of $E$
that is specific to $(s,p,t_k)$ (namely a \emph{historical vocabulary} for $(s,p,t_k)$) as $\mathbf{H}_{t_k}^{(s,p)}$, which contains all the entities that have served as object entities in facts with the subject entity $s$ and the predicate $p$ along the known snapshots $\mathcal{G}_{(t_1,t_{k-1})}=\{\mathcal{G}_1,\mathcal{G}_2,...,\mathcal{G}_{k-1}\}$ before $t_k$, where the historical vocabulary $\mathbf{H}_{t_k}^{(s,p)}$ is an $N$-dimensional multi-hot indicator vector and $N$ is the cardinality of $\mathcal{E}$, the value of entities in the historical vocabulary are marked $1$ while others are $0$. 
Prediction of a missing temporal fact aims to infer the missing object entity given $(s,p,?,t)$ (or the subject entity given $(?,p,o,t)$, or the predicate with subject entity and object entity given $(s,?,o,t)$).
Without loss of generality, we describe our model as predicting the missing object entity in a temporal fact, although the model can be easily extended to predicting other elements including the subject entity and the predicate.

\nop{
TKGs are composed of quadruples $(s,p,o,t)$ \muhao{this is not a set. a set is $\{.\}$} or $(s_t,p_t,o_t)$ \muhao{the latter is not a quadruple. why do you need it anyway?} modeling facts at specific times. Each of the fact captures the relation (or predicate) $p$ for subject and object entities $s\in \mathcal{E}$ and $o\in \mathcal{E}$ at time step $t\in T$, 
$\cS=\{s_1,s_2,…,s_N\}$, $\cP=\{p_1,p_2,…,p_M\}$ \muhao{Redflag: are you sure you need two vocabularies of entities?}, and $\mathcal{T}=\{t_1,t_2,…,t_T\}$ respectively denote the corresponding vocabularies of entities, predicate and timestamp \muhao{you cannot have a ``vocabulary'' of time.}.
\muhao{I would replace the previous sentence as (check the rest of the paper to make sure notations are consistent):} where $\mathcal{E}$ is the vocabulary of entities and $\mathcal{T}$ is the set of timestamps.
$\mathcal{G}_t$ is a snapshot of KG at at time $t$. $g=(s,p,o,t) \in \mathcal{G}_t$ denotes a quadruple such that $s,o\in \cS$, $p\in \cP$ and $t\in T$. A TKG is built upon a sequence of fact quadruples ordered ascending based on their timestamps, i.e., $\mathcal{G}=\{\mathcal{G}_1,\mathcal{G}_2,…,\mathcal{G}_T\}$ \muhao{You have defining $\mathcal{G}$ as a set instead of sequence now. If it is a sequence than should be $[.]$ instead of $\{.\}$}. 
\muhao{where ... is removed for redundancy.}
There will be a phenomenon that the same triple $(s,r,o)$ may appear multiple times in different timestamps, yielding different fact quadruples \muhao{This sentence: 1. what is a triple? 2. This phenomenon does not belong to notations (you are not formalizing it). So suggest discarding it.}.
For each snapshot at time step $t_k$, the known vocabulary $\mathbf{H}_{t_k}\in \mathbb{R}^{(N\times M)\times N}$ is a multi-hot indicator vector which includes all object entities that occur under any subject entity and predicate before time step $t_k$.
\muhao{You have not explained what does a ``known vocabulary'' mean conceptually. This is also a confusing term, since the entire entity vocabulary is already known to you.}
For example, $\mathbf{H}_{t_k}^{(s_i,p_j)}$ is an N-dimensional multi-hot indicator vector for the $i\times j$th row of $\mathbf{H}_{t_k}$, where the value of the object entities on the known facts is $1$, others are $0$. 
}

\subsection{Model Components}

As shown in Figure~\ref{fig:model}, our model combines two modes of inference, namely \emph{Copy mode} and \emph{Generation mode},
where the former seeks to select entities from a specific historical vocabulary that forms repeated facts in history while the latter predicts entities from the whole entity vocabulary.
When predicting a quadruple $(s_1,p_3,?,T+1)$ (as shown in Figure~\ref{fig:model}), the Copy mode should infer the probability of entities in the historical vocabulary $\{s_3,s_4,s_5,...,s_m\}$ which have served as the object entities in facts of the subject entity $s_1$ and the predicate $p_3$ along the known snapshots $\mathcal{G}_{(t_1,t_T)}$. On the other hand, the Generation mode estimates the probability of every entity in the whole entity vocabulary to answer a query. Then \model combines the probabilistic predictions from both Copy mode and Generation mode to output the final prediction.

\begin{figure}[!t]
\centering
\includegraphics[width=0.47\textwidth]{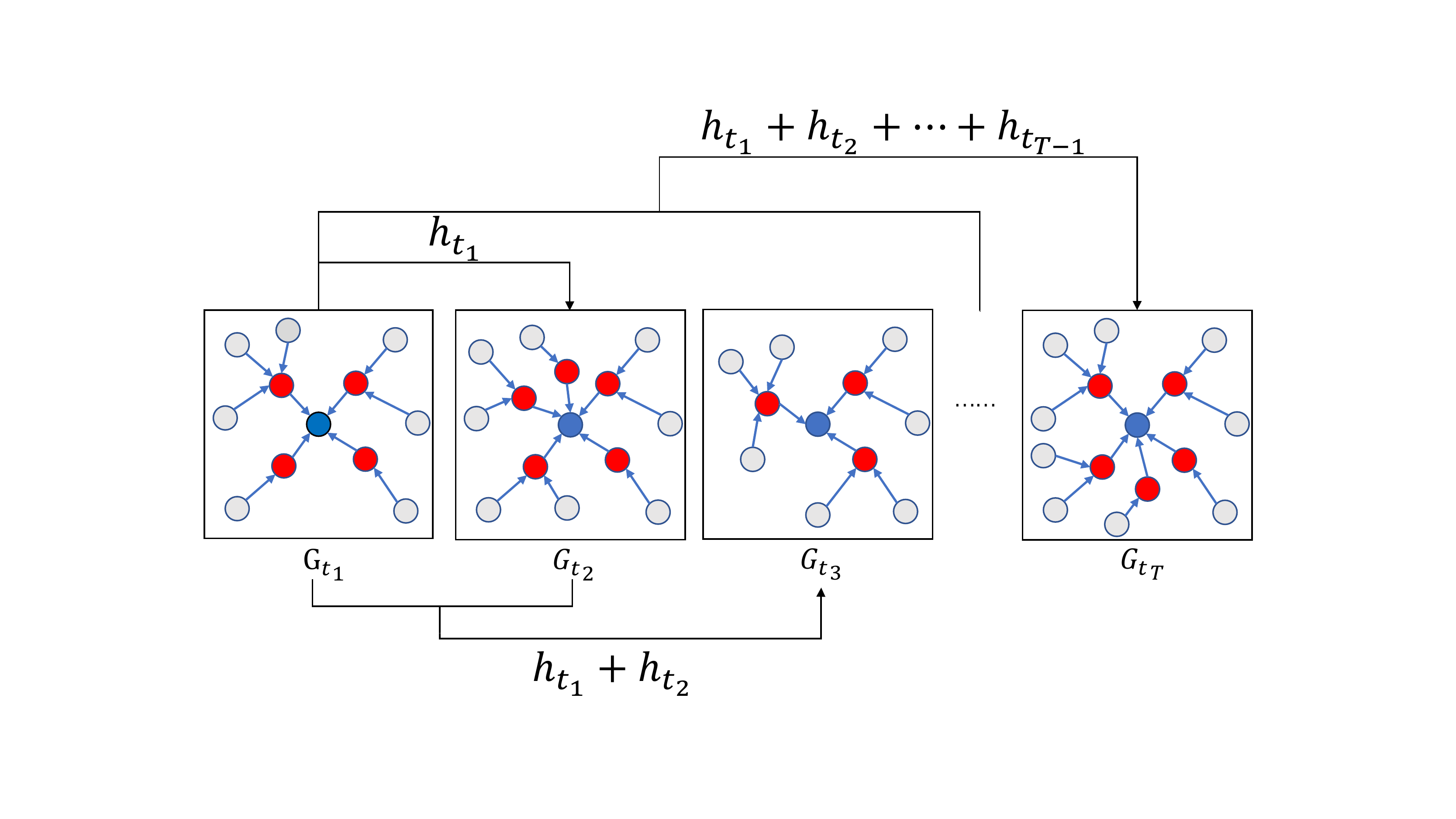}
\caption{Illustration of the training process. Each time we train on a new TKG snapshot, we will extend the historical vocabulary based on that of the previous snapshot. $h_{t_k}$ contains the historical vocabulary of all subject entity and predicate pair in the snapshot $G_{t_k}$.} \label{fig:training}
\end{figure}

We first process the training set to obtain the historical vocabulary for each subject entity and predicate 
combination $(s,p,t)$ on every $t$ in training snapshots, i.e. $\{\mathbf{h}_{t_1}^{(s,p)},\mathbf{h}_{t_2}^{(s,p)},...,\mathbf{h}_{t_T}^{(s,p)}\}$, where $\mathbf{h}_{t_k}^{(s,p)}$ is an $N$-dimensional multi-hot indicator vector that includes all object entities served as object entities in facts with the subject $s$ and the predicate $p$ in snapshot $\mathcal{G}_{t_k}$. 
As shown in Figure~\ref{fig:training}, 
we train the model sequentially on each snapshot, similar to the idea of recursion, trained by incrementally maintain the historical vocabulary for all the previous snapshots.
When we evaluate \model's performances in the validation set and test set, the maximum historical vocabulary from the whole training set will be used.


Specifically,
for each query quadruple $(s,p,?,t_k)$ on time $t_k$, the training process extends the historical vocabulary that specific to $(s,p,t_k)$ from that of the previous snapshot, as formalized below:

\begin{equation}
    \mathbf{H}_{t_k}^{(s,p)}=\mathbf{h}_{t_1}^{(s,p)}+\mathbf{h}_{t_2}^{(s,p)}+...+\mathbf{h}_{t_{k-1}}^{(s,p)},
\end{equation}
where $\mathbf{H}_{t_k}^{(s,p)}$ is an $N$-dimensional multi-hot indicator vector where $1$ is marked for all entities in the current historical vocabulary. We now introduce the two inference modes in the following.


\stitle{Copy Mode}
The Copy mode is designed to identify facts with recurrence, and accordingly predict the future facts by copying from known facts in history.

If query $(s,p,?,t_k)$ has the historical vocabulary $\mathbf{H}_{t_k}^{(s,p)}$ specific to the subject entity $s$ and the predicate $p$ at time step $t_k$,
\model will increase the probability estimated for the object entities that are selected in the historical vocabulary. In detail, the Copy mode first generates an index vector $v_{q}$ with an MLP:

\begin{equation}
    \mathbf{t_k}=\mathbf{t_{k-1}}+\mathbf{t_u},
\end{equation}
\begin{equation}
    \mathbf{v}_q=\mathrm{tanh}(\mathbf{W}_c[\mathbf{s},\mathbf{p},\mathbf{t_k}]+\mathbf{b}_c) ,
\end{equation}
where $\mathbf{W}_c \in \mathbb{R}^{3d\times N}$ and $\mathbf{b}_c \in \mathbb{R}^{N}$ are trainable parameters. $\mathbf{t_u}$ is a unit step of time and $\mathbf{t_1}=\mathbf{t_u}$. The index vector $\mathbf{v}_q$ is an $N$-dimensional vector, where $N$ is the cardinality of the whole entity vocabulary $\mathcal{E}$. 

To minimize the probability of some entities that do not form known facts with $s$ and $p$ in history (i.e. those being \emph{uninterested} to the Copy mode), we first make modifications to $\mathbf{H}_{t_k}^{(s,p)}$. $\mathbf{\dot{H}}_{t_k}^{(s,p)}$ changes the index value for an uninterested entity in $\mathbf{H}_{t_k}^{(s,p)}$ to a small negative number.
Therefore, \model can delimit the candidate space by adding the index vector $\mathbf{v}_q$ and the changed multi-hot indicator vector $\mathbf{\dot{H}}_{t_k}^{(s,p)}$,
minimize the probability of the uninterested entities, and then estimate the probability of the object entities in the historical vocabulary with a softmax function:

\begin{equation}
    \mathbf{c}_q = \mathbf{v}_q+\mathbf{\dot{H}}_{t_k}^{(s,p)},
\end{equation}
\begin{equation}
    \mathbf{p}(c)=\mathrm{softmax}(\mathbf{c}_q),
\end{equation}
where $\mathbf{c}_q$ is an $N$-dimensional index vector, such that the values corresponding to \emph{uninterested} entities in $\mathbf{c}_q$ are assigned close to zero. $\mathbf{p}(c)$ is a vector whose size equals to that of the whole entity vocabulary, and is to represent the prediction probabilities on the historical vocabulary. 
Eventually, the largest dimension of $\mathbf{p}(c)$ indicates the object entity to be copied from the historical vocabulary. The merit of the Copy mode is that it is able to learn to predict from a much delimited candidate space than the entire entity vocabulary. However, facts can also be newly appearing in a upcoming snapshot. And we therefore need a Generation mode to predict such facts.

\stitle{Generation Mode} Given the same aforementioned query $(s,p,?,t_k)$, the
Generation mode is responsible for predicting facts by selecting the object entity from the whole entity vocabulary $\mathcal{E}$.
The prediction made by the generation mode regards the predicted fact as an entirely new fact without any references to history.
Similar to the Copy mode, the Generation mode also generates an index vector $\mathbf{g}_q$ whose size equals to the size of the candidate space $\mathcal{E}$, 
and 
is normalized with a softmax function to make predictions:

\begin{equation}
    \mathbf{g}_q=\mathbf{W}_g[\mathbf{s},\mathbf{p},\mathbf{t_k}]+\mathbf{b}_g,
\end{equation}
\begin{equation}
    \mathbf{p}(g)=\mathrm{softmax}(\mathbf{g}_q),
\end{equation}
where $\mathbf{W}_g\in \mathbb{R}^{3d\times N}$ and $\mathbf{b}_g\in \mathbb{R}^{N}$ are trainable parameters.
Similar to $\mathbf{p}(c)$ in the Copy mode, $\mathbf{p}(g)$ 
represents the predicted probability on the whole entity vocabulary. The maximum value in $\mathbf{p}(g)$ indicates the object entity we predict in the whole entity vocabulary by the Generation mode. The Generation mode complements the Copy mode with the ability for \emph{de novo} fact prediction.

\subsection{Learning Objective}\label{sec:loss_function}

Predicting the (object) entity when given a query $(s,p,?,t)$ can be viewed as a multi-class classification task, where each class corresponds to one object entity. 
The learning objective is to minimize the following cross-entropy loss $\mathcal{L}$ on all facts of the TKG snapshots that exist during training: 

\begin{equation}
    \mathcal{L}=-\sum_{t\in T}\sum_{i\in \mathcal{E}}\sum_{k=1}^{K}{o_{it}\ln \mathbf{p}(y_{ik}|s,p,t)}
\end{equation}
where $o_{it}$ is the $i$-th ground truth object entity in the snapshot $\mathcal{G}_t$, $\mathbf{p}(y_{ik}|s,p,t)$ is the combined probability value of the $k$-th object entity in the snapshot $\mathcal{G}_t$ when the $i$-th ground truth object entity is $o_i$.

\subsection{Inference}
Without loss of generality, we describe the inference process as predicting the missing object in a temporal fact, although this process can be easily extended to predicting the subject and the relation as well.
To make prediction regarding a query $(s,p,?,t_k)$, 
both Copy and Generation modes give a predicted object entity with the highest probability within their candidate spaces, whereof the Copy mode may make predictions from a much smaller candidate space than the entire entity vocabulary. 
To ensure that the sum of the probability equals $1$ for all entities in $\mathcal{E}$, a coefficient $\alpha$ is incorporated to adjust the weight between the Copy mode and the Generation mode.
\model combines the probabilistic predictions from both the Copy mode and the Generation mode by adding the probability of each entity given by these two modes. 
The final prediction $o_t$ will be the entity that receives the highest combined probability, as defined below: 

\begin{equation}
    \mathbf{p}(o|s,p,t)=\alpha*\mathbf{p}(c)+(1-\alpha)*\mathbf{p}(g),
\end{equation}
\begin{equation}
    o_t=\mathrm{argmax}_{o\in \mathcal{E}}\mathbf{p}(o|s,p,t),
\end{equation}
where $\alpha \in [0,1]$, $\mathbf{p}(o|s,p,t)$ is the whole entity vocabulary size vector which contains the probability of all entities.

\nop{
Integrating the probability \muhao{What do you mean by integrating (it is not a right word to use anyway) the probability? I bet you are not even able to translate into Chinese. Do you select one of the two predictions with a higher probability?} of Copy mode and generation mode reasonably might improve the individual performance of both modules greatly that means it is capability to balance the utilization between the known vocabulary and the whole entity vocabulary. The probability of generating an target object entity $y_t$ is given by the ''mixture'' \muhao{what is the CHINESE meaning of mixture her?} of probability as follows:

\begin{equation}
    \mathbf{p}(y_t)=\alpha*\mathbf{p}(c)+(1-\alpha)*\mathbf{p}(g)
\end{equation}
where $\alpha \in (0,1)$ is a hyper-parameter set empirically. The maximum value in probability $\mathbf{p}(y_t)$ is the target object entity $y_t$ we predict. 
}
\section{Experiments}
\label{sec:experimental}
In this section, we demonstrate the effectiveness of \model with five public TKG datasets. We first explain experimental settings in detail, including details about baselines and datasets. After that, we discuss the experimental results. We also conduct an ablation study to evaluate the importance of different components of \model\footnote{The released source code and documentation are available at \url{https://github.com/CunchaoZ/CyGNet}.}.

\begin{table*}[!t]
\centering
{\footnotesize
\begin{tabular}{cccccccc}
\toprule[2pt]
\#Data  & \#Entities  & \#Relation & \#Training & \#Validation & \#Test & \#Granularity & \#Time Granules \\ \midrule[1pt]
ICEWS18 & 23,033 & 256 & 373,018 & 45,995 & 49,545 & 24 hours & 304 \\
ICEWS14 & 12,498 & 260 & 323,895 & - & 341,409 & 24 hours   & 365 \\
GDELT & 7,691 & 240 & 1,734,399 & 238,765 & 305,241 & 15 mins & 2,751 \\
WIKI & 12,554 & 24 & 539,286 & 67,538 & 63,110 & 1 year     & 232 \\ 
YAGO & 10,623 & 10 & 161,540 & 19,523 & 20,026 & 1 year     & 189 \\\bottomrule[2pt]
\end{tabular}
}
\caption{Statistics of the datasets.}\label{tbl:stat}
\end{table*}

\begin{table*}[!ht]
\footnotesize
\centering
\scalebox{0.9}{
\begin{tabular}{ccccccccccccc}
\toprule[2pt]
\multirow{2}{*}{Method} & \multicolumn{4}{c}{ICEWS18} & \multicolumn{4}{c}{ICEWS14} & \multicolumn{4}{c}{GDELT} \\ \cline{2-13} 
 & MRR & Hits@1 & Hits@3 & Hits@10 & MRR & Hits@1 & Hits@3 & Hits@10 & MRR & Hits@1 & Hits@3 & Hits@10 \\ \midrule[1pt]
TransE & 17.56 & 2.48 & 26.95 & 43.87 & 18.65 & 1.12 & 31.34 & 47.07 & 16.05 & 0.00 & 26.10 & 42.29 \\
DistMult & 22.16 & 12.13 & 26.00 & 42.18 & 19.06 & 10.09 & 22.00 & 36.41 & 18.71 & 11.59 & 20.05 & 32.55 \\
ComplEx & 30.09 & 21.88 & 34.15 & 45.96 & 24.47 & 16.13 & 27.49 & 41.09 & 22.77 & 15.77 & 24.05 & 36.33 \\
R-GCN & 23.19 & 16.36 & 25.34 & 36.48 & 26.31 & 18.23 & 30.43 & 45.34 & 23.31 & 17.24 & 24.96 & 34.36 \\
ConvE & 36.67 & 28.51 & 39.80 & 50.69 & 40.73 & 33.20 & 43.92 & 54.35 & 35.99 & 27.05 & 39.32 & 49.44 \\
RotatE & 23.10 & 14.33 & 27.61 & 38.72 & 29.56 & 22.14 & 32.92 & 42.68 & 22.33 & 16.68 & 23.89 & 32.29 \\ \hline
HyTE & 7.31 & 3.10 & 7.50 & 14.95 & 11.48 & 5.64 & 13.04 & 22.51 & 6.37 & 0.00 & 6.72 & 18.63 \\
TTransE & 8.36 & 1.94 & 8.71 & 21.93 & 6.35 & 1.23 & 5.80 & 16.65 & 5.52 & 0.47 & 5.01 & 15.27 \\
TA-DistMult & 28.53 & 20.30 & 31.57 & 44.96 & 20.78 & 13.43 & 22.80 & 35.26 & 29.35 & 22.11 & 31.56 & 41.39 \\
Know-Evolve+MLP & 9.29 & 5.11 & 9.62 & 17.18 & 22.89 & 14.31 & 26.68 & 38.57 & 22.78 & 15.40 & 25.49 & 35.41  \\
DyRep+MLP & 9.86 & 5.14 & 10.66 & 18.66 & 24.61 & 15.88 & 28.87 & 39.34 & 23.94 & 15.57 & 27.88 & 36.58 \\
R-GCRN+MLP & 35.12 & 27.19 & 38.26 & 50.49 & 36.77 & 28.63 & 40.15 & 52.33 & 37.29 & 29.00 & 41.08 & 51.88 \\
RE-NET & \underline{42.93} & \underline{36.19} & \underline{45.47} & \underline{55.80} & \underline{45.71} & \underline{38.42} & \underline{49.06} & \underline{59.12} & \underline{40.12} & \underline{32.43} & \underline{43.40} & \underline{53.80} \\ \hline
\model & \pmb{46.69} & \pmb{40.58} & \pmb{49.82} & \pmb{57.14} & \pmb{48.63} & \pmb{41.77} & \pmb{52.50} & \pmb{60.29} & \pmb{50.92} & \pmb{44.53} & \pmb{54.69} & \pmb{60.99} \\ \bottomrule[2pt]
\end{tabular}}
\caption{Results (in percentage) on three datasets. The best results are boldfaced, and the second best ones are underlined.}\label{tab:results on ICEWS}
\end{table*}

\subsection{Experimental Setup}
\label{sec:set_up}

\stitle{Datasets} 
We evaluate \model on the task of link prediction using five benchmark datasets, namely ICEWS18, ICEWS14, GDELT, WIKI and YAGO specifically. ICEWS records political 
facts (events) with timestamps, e.g., $($\emph{Donald~Trump}, \emph{visit}, \emph{France}, 2018-04-10$)$, and there are 
two benchmark datasets extracted from two time periods in this knowledge base 
i.e., ICEWS18 (\cite{boschee2015icews}; from 1/1/2018 to 10/31/2018) and ICEWS14 (\cite{trivedi2017know}; from 1/1/2014 to 12/31/2014).
GDELT \citep{leetaru2013gdelt} 
is a catalog of human societal-scale behavior and beliefs extracted from news media, and the experimental dataset
is collected from 1/1/2018 to 1/31/2018 with a time granularity of 15 mins. The WIKI and YAGO datasets 
are subsets of the Wikipedia history and YAGO3 \citep{mahdisoltani2013yago3}, respectively.
Since the WIKI and YAGO datasets contain temporal facts with a time span in the form as $(s,p,o,[T_s,T_e])$, where $T_s$ is the starting time and $T_e$ is the ending time, we follow prior work \citep{jin2020Renet} to discretize these temporal facts into snapshots with the time granularity of a year. Table~\ref{tbl:stat} summarizes the statistics of these datasets.

\stitle{Baseline Methods} 
We compare our method with a wide selection of static KGE and TKGE models. 
Static KGE methods include TransE \citep{bordes2013translating}, DistMult \citep{yang2014embedding}, ComplEX \citep{trouillon2016complex}, R-GCN \citep{schlichtkrull2018modeling}, ConvE \citep{dettmers2018convolutional} and RotatE \citep{sun2019rotate}. 
Temporal methods include TTransE \citep{jiang2016towards}, HyTE \citep{dasgupta2018hyte}, TA-DistMult \citep{garcia2018learning},
Know-Evolve+MLP \citep{trivedi2017know}, DyRep+MLP \citep{trivedi2018dyrep}, R-GCRN+MLP \citep{seo2018structured}, and RE-NET \citep{jin2020Renet}, whereof RE-NET has offered SOTA performance on all of the benchmarks. Know-Evolve+MLP, DyRep+MLP, R-GCRN+MLP are the former methods in combination with MLP decoder.
More detailed descriptions of baseline methods are given in Appendix~\ref{sup:baseline} \cite{zhu2021learning}.

\stitle{Evaluation Protocols}
Following prior work \cite{jin2020Renet},
we split each dataset except for ICEWS14 into training set, validation set and testing set into $80$\%$/10$\%$/10$\% splits in the chronological order, 
whereas ICEWS14 is not provided with a validation set.
We report Mean Reciprocal Ranks (MRR) and Hits@1/3/10 (the proportion of correct test cases that are ranked within top 1/3/10). We also enforce the \emph{filtered} evaluation constraint that has been widely adopted in prior work \citep{jin2020Renet,dasgupta2018hyte}.



\stitle{Model Configurations}
The values of hyper-parameters are determined according to the MRR performance on each validation set. 
Specifically, since ICEWS14 does not come with a validation set, we directly carry forward the same hyper-parameter settings from ICEWS18 to ICEWS14. 
The coefficient $\alpha$ is tuned from 0.1 to 0.9 with a step of 0.1.
It is set to 0.8 for ICEWS18 and ICEWS14, 0.7 for GDELT, and 0.5 for YAGO and WIKI. 
The model parameters are first initialized with Xavier initialization \citep{glorot2010understanding},
and are optimized using an AMSGrad optimizer with a learning rate of 0.001. 
The batch size is set to 1024.
The training epoch is limited to 30, which is enough for convergence in most cases. 
The embedding dimension is set to 200 to be consistent with the setting by \citet{jin2020Renet}. 
The baseline results are also adopted from \citet{jin2020Renet}.

\begin{table}[t]
\setlength{\tabcolsep}{2pt}
\footnotesize
\centering
\scalebox{0.9}{
\begin{tabular}{ccccccc}
\toprule[2pt]
\multirow{2}{*}{Method} & \multicolumn{3}{c}{WIKI}  & \multicolumn{3}{c}{YAGO} \\ \cline{2-7} 
 & MRR & Hits@3 & Hits@10 & MRR & Hits@3 & Hits@10 \\ \midrule[1pt]
TransE & 46.68 & 49.71 & 51.71 & 48.97 & 62.45 & 66.05\\
DistMult & 46.12 & 49.81 & 51.38 & 59.47 & 60.91 & 65.26\\
ComplEx & 47.84 & 50.08 & 51.39 & 61.29 & 62.28 & 66.82\\
R-GCN & 37.57 & 39.66 & 41.90 & 41.30 & 44.44 & 52.68\\
ConvE & 47.57 & 50.10 & 50.53 & 62.32 & 63.97 & 65.60\\
RotatE & \underline{50.67} & 50.71 & 50.88 & \underline{65.09} & 65.67 & 66.16\\ \hline
HyTE & 43.02 & 45.12 & 49.49 & 23.16 & 45.74 & 51.94\\
TTransE & 31.74 & 36.25 & 43.45 & 32.57 & 43.39 & 53.37\\
TA-DistMult & 48.09 & 49.51 & 51.70 & 61.72 & 65.32 & 67.19\\
Know-Evolve+MLP & 12.64 & 14.33 & 21.57 & 6.19 & 6.59 & 11.48\\
DyRep+MLP & 11.60 & 12.74 & 21.65 & 5.87 & 6.54 & 11.98\\
R-GCRN+MLP & 47.71 & 48.14 & 49.66 & 53.89 & 56.06 & 61.19\\
RE-NET & \pmb{51.97} & \pmb{52.07} & \pmb{53.91} & \pmb{65.16} & 65.63 & 68.08\\ \hline
\model & 45.50 & \underline{50.79} & \underline{52.80} & 63.47 & \pmb{65.71} & \pmb{68.95}\\ \bottomrule[2pt]
\end{tabular}}
\caption{Results (in percentage) on WIKI and YAGO. The best results are boldfaced, and the second best are underlined. Hits@1 was not reported by \citet{jin2020Renet} on these datasets, for which \model achieved 41.73\% on WIKI and 60.04\% on YAGO.
} \label{tab:results on YAGO and WIKI}
\end{table}


\begin{table}[t]
\setlength{\tabcolsep}{5pt}
\footnotesize
\centering
\scalebox{0.9}{
\begin{tabular}{ccccccc}
\toprule[2pt]
\multirow{2}{*}{Method} & \multicolumn{3}{c}{Predicting subjects} & \multicolumn{3}{c}{Predicting objects} \\ \cline{2-7} 
 & MRR & Hits@3 & Hits@10 & MRR & Hits@3 & Hits@10 \\ \midrule[1pt]
RE-NET & \pmb{50.44} & \pmb{50.80} & \pmb{52.06} & \pmb{53.48} & 53.47 & 55.63 \\  
\model & 37.80 & 42.20 & 49.97 & 53.04 & \pmb{53.85} & \pmb{55.91} \\ \bottomrule[2pt]
\end{tabular}}
\caption{Results (in percentage) for predicting subject or object entities separately on WIKI.}
\label{tab:why WIKI}
\end{table}

\subsection{Results}
\label{sec:results}

Tabels~\ref{tab:results on ICEWS} and~\ref{tab:results on YAGO and WIKI} report the link prediction results by \model and baseline methods on five TKG datasets. 
On the first three datasets, static KGE methods fell far behind RE-NET or \model due to their inability to capture temporal dynamics.
It can also be observed that all static KGE methods generally perform better than TTransE and HyTE. 
We believe this is due to that TTransE and HyTE only learn representations independently for each snapshot, while lack capturing the long-term dependency.
Table~\ref{tab:results on ICEWS} also shows that \model significantly outperforms other baselines on ICEWS18, ICEWS14 and GDELT.
For instance, \model achieves the improvements of 10.80\% (MRR), 12.10\% (Hits@1), 11.29\% (Hits@3), and 7.19\% (Hits@10) over state-of-the-arts on GDELT. We later found that GDELT has denser training facts in each snapshot, indicating it has more historical information that \model could utilize to improve the prediction accuracy.


We also observe in Table~\ref{tab:results on YAGO and WIKI} that \model does not always perform the best on WIKI and YAGO, especially on WIKI. To give further analyses, we separate the results for predicting subject entities and those for object entities (Table~\ref{tab:why WIKI}). We can observe that \model performs much better in predicting object entities on WIKI. 
This is in fact due to the unbalanced proportions of subjects and objects in participating recurring facts on WIKI. More specifically, if we group the facts by subjects and relations, we observe as many as 62.3\% of the groups where the objects have repeated for at least once in history, whereas the recurrence rate of subject entities (in facts grouped by relations and objects) is around 23.4\%.
The imbalanced recurrence rate has hindered the combined learning of \model's two components, hence caused a deteriorated effect to its overall prediction performance.
While \model performs much better on the other datasets with more balanced recurrence rates of subject and object entities, how to tackle this shortcoming of \model with a more robust meta-learning framework is a meaningful direction for further studies.

\subsection{Ablation Study}
\label{sec:ablation_study}


\begin{table}[!t]
\centering
{\footnotesize
\scalebox{0.94}{
\setlength{\tabcolsep}{1mm}{
\begin{tabular}{ccccc}
\toprule[2pt]
\multirow{2}{*}{Method}  & \multicolumn{4}{c}{ICEWS18} \\ \cline{2-5} 
 & MRR & Hits@1 & Hits@3 & Hits@10 \\ \midrule[1pt]
\model-Copy-only & 42.60 & 38.47 & 44.93 & 49.00 \\
\model-Generation-only & 34.58 & 25.94 & 38.39 & 51.05 \\ 
\model-Generation-new & 43.88 & 37.17 & 47.03 & 56.25 \\ 
\hline
\model & \pmb{46.69} & \pmb{40.58} & \pmb{49.82} & \pmb{57.14} \\ \bottomrule[2pt]
\end{tabular}
}}}
\caption{Results (in percentage) by different variants of our model on ICEWS18.}\label{tab:variant results}
\end{table}

To help understand the contribution of different model components of \model, we present an ablation study. To do so, we create variants of \model by adjusting the use of its model components, and compare the performance on the ICEWS18 dataset. 
Besides, more analyses are given in Appendix~\ref{sup:hyper} \cite{zhu2021learning} due to space limitation.

From the results in Tables~\ref{tab:variant results}, we observe that the Copy mode and the Generation mode are both important. Removing the Copy mode can lead to a drop of 12.11\% in MRR, as well as drastic drops of other metrics, which shows that learning to predict future facts by referring to the known facts in the past can be helpful. 
On the other hand, the removal of the Generation mode leads to a drop of 4.09\% in MRR, 
which counts for the loss of the model's ability for \emph{de novo} fact prediction.
These results further explain that the promising performance by \model is due to both the capability of learning from history, and identifying and predicting new facts from scratch.

\model-Generation-new is another variant of \model. The difference between \model-Generation-new and the complete \model is that the former utilizes the Generation mode to predict new facts in the whole entity vocabulary modulo the historical vocabulary. The performance of \model is noticeably better than \model-Generation-new. 
We believe this is because the original Generation mode in \model can also strengthen the prediction in cases where a future fact is repeated.
This merit is however discarded in the modified Generation mode of \model-Generation-new.

\section{Conclusion}\label{sec:conclusion}

Characterizing and inferring temporal knowledge is a challenging problem. In this paper, 
we leverage the copy mechanism to tackle this problem, based on the hypothesis that a future fact can be predicted from the facts in history. The proposed \model is not only able to predict facts from the whole open world, but is also capable of identifying facts with recurrence and accordingly selecting such future facts based on the known facts appeared in the past. The presented results demonstrate \model's promising performance for predicting future facts in TKGs.
For future work, 
we plan to improve the sequential copy mechanism 
by identifying globally salient entities and events \cite{fan2019learning} in TKGs.
Grounding dated documents to TKGs with the obtained embedding representations is also a meaningful direction.
Besides, we are interested in leveraging the proposed techniques to help understand dynamic event processes in natural language text \cite{chen-etal-2020-what,zhang-etal-2020-analogous}.

\section*{Acknowledgment}

We appreciate the anonymous reviewers for their insightful comments.
This work is partially supported by NSFC-71701205, NSFC-62073333.
Muhao Chen's work is supported by Air Force Research Laboratory under agreement number FA8750-20-2-10002, and by the DARPA MCS program under Contract No. N660011924033.

\bibstyle{aaai21}
\bibliography{ref}

\clearpage
\appendix

\begin{center}
\LARGE
    \textbf{Supplementary Material}
\end{center}

\section{Hyper-parameter Analysis}\label{sup:hyper}

\begin{figure}[htbp]
\centering
\includegraphics[width=0.48\textwidth]{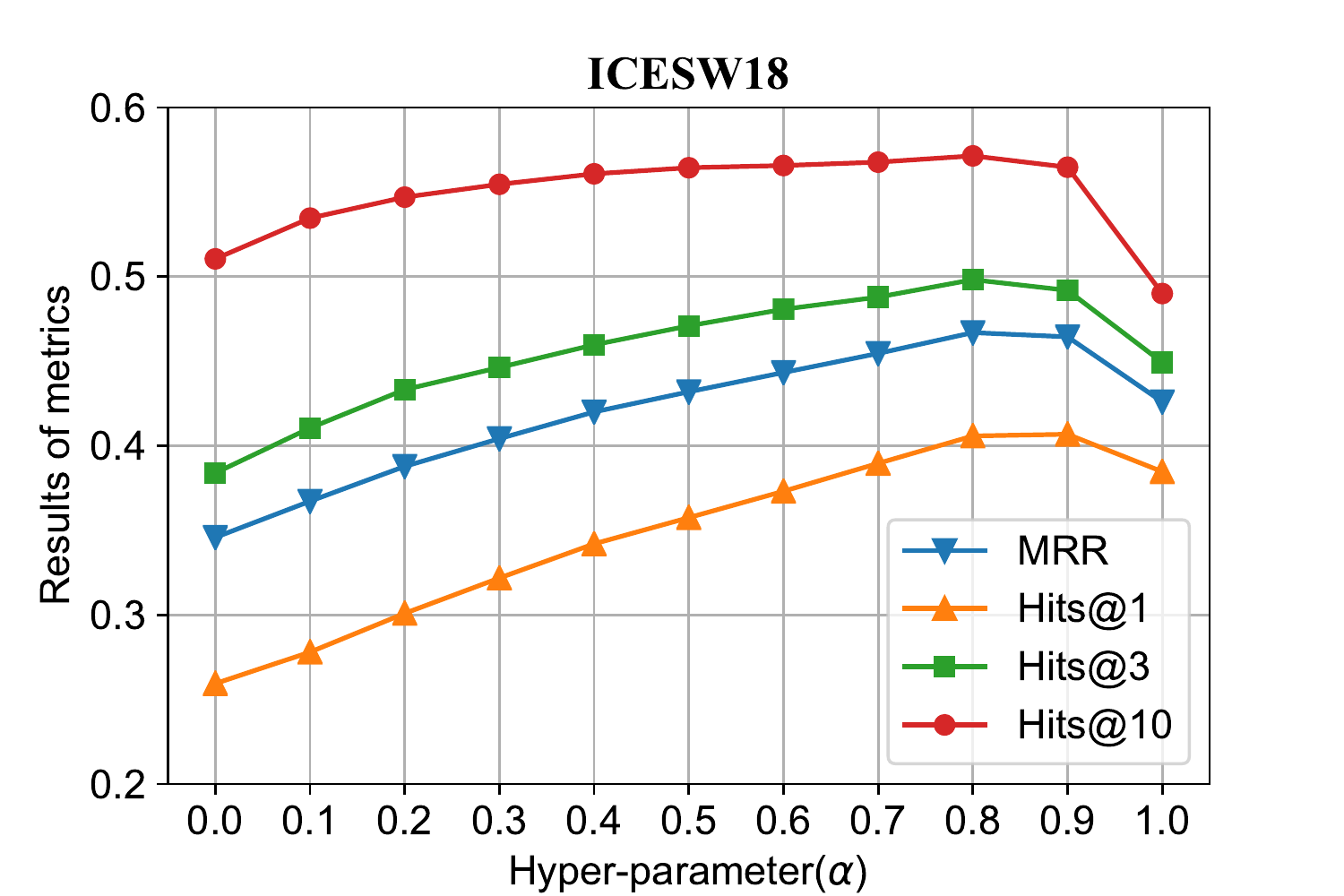}
\caption{Results for different hyper-parameter $\alpha$ of \model on ICEWS18.}\label{fig:results for different hyper-parameter}
\end{figure}

To help understand the contribution of different model components in \model, we adjust the coefficient $\alpha$ to change the weight between the Copy mode and the Generation mode. The results are presented in Figure~\ref{fig:results for different hyper-parameter}.

The hyper-parameter $\alpha$ is $0$ means that \model only uses the Generation mode and $\alpha$ is $1$ means that \model only uses the Copy mode. 
As we can observe that, without considering the known facts which occurred in the past ($\alpha=0$), the performance of \model that only uses the Generation mode is not sufficiently effective.
And within a certain range, with the increase of $\alpha$, the performance of \model increases. 
Excessive consideration of the known facts can lead to decrease the performance of \model. And the most extreme is that \model only uses the Copy mode that ignores the \emph{de nove} facts, i.e., $\alpha=1$. 
Thus we assume that \model has the capability to balance the utilization between the historical vocabulary and the whole entity vocabulary by adjusting the coefficient $\alpha$.

\section{Descriptions of Baseline Methods}\label{sup:baseline}

We provide descriptions of baseline methods. In accord with Section~\ref{sec:set_up}, we separate the description in two groups.

TransE \citep{bordes2013translating} represents entities and relations in $d$-dimension vector space and makes the added embedding of the subject entity $s$ and the predicate $p$ be close to the embedding of the object entity $o$, i.e., $s+p\approx o$.
DistMult \citep{yang2014embedding} a simplified bilinear model by restricting relation matrix to be diagonal matrix for multi-relational representation learning.
ComplEx \citep{trouillon2016complex} firstly introduces complex vector space which can capture both symmetric and antisymmetric relations.
Relational Graph Convolutional Networks (RGCN;\citet{schlichtkrull2018modeling}) applies the previous Graph Convolutional Networks (GCN;\citet{kipf2017semi}) works to relational data modeling.
ConvE \citep{dettmers2018convolutional} uses $2D$ convolution over embeddings and multiple layers of nonlinear features, and reshapes subject entity and predicate into $2D$ matrix to model the interactions between entities and relations.
RotatE \citep{sun2019rotate} proposes a rotational model taking predicate as a rotation from subject entity to object entity in complex sapce as $o=s\circ p$ where $\circ$ denotes the element-wise Hadmard product.
However, these static KGE methods do not capture temporal facts.

More recent attempts have been made to model the evolving facts in TKGs. 
TTransE \citep{jiang2016towards} is an extension of TransE by embedding temporal information into the score function. 
HyTE \citep{dasgupta2018hyte} replaces the projection normal vector in TransH \citep{wang2014knowledge} with a time-related normal vector. 
Know-Evolve \citep{trivedi2017know} learns non-linearly evolving entity representations over time which models the occurrence of a fact as a temporal point process. 
TA-DistMult \citep{garcia2018learning} utilize recurrent neural networks to learn time-aware representations of relations and uses standard scoring functions from TransE and DistMult. 
RE-NET \citep{jin2020Renet} models event (fact) sequences via RNN-based event encoder and neighborhood aggregator.
DyREP \citep{trivedi2018dyrep} divides the dynamic graph network into two processes, and uses representation learning as a potential bridge connecting the two processes to learn the temporal structure information in the network. 
GCRN \citep{seo2018structured} merges CNN for graph-structured data and RNN to identify simultaneously meaningful spatial structures and dynamic patterns. 
Know-evolve, DyREP and GCRN combined with MLP decoder to predict future facts, which are called Know-evolve+MLP, DyRep+MLP and R-GCNR+MLP in \citep{jin2020Renet}. 


\end{document}